\documentclass[a4paper,twoside]{article}

\usepackage{epsfig}
\usepackage{graphicx}
\usepackage{subcaption}
\usepackage{calc}
\usepackage{amssymb}
\usepackage{comment}
\usepackage{amstext}
\usepackage{amsmath}
\usepackage{amsthm,amssymb}
\usepackage{color}
\usepackage{cite}
\usepackage{algorithm}
\usepackage{algpseudocode}
\usepackage{verbatim}
\usepackage{multirow}
\usepackage{multicol}
\usepackage{setspace}
\usepackage{bbm}
\usepackage{pslatex}
\usepackage{apalike}
\usepackage[bottom]{footmisc}
\usepackage{SCITEPRESS}     

\begin{document}

\title{Robust Semi-Supervised Anomaly Detection via \\ Adversarially Learned Continuous Noise Corruption}

 \author{
 \authorname{Jack W. Barker\sup{1}, Neelanjan Bhowmik\sup{1}, Yona Falinie A. Gaus\sup{1} and Toby P. Breckon\sup{1, 2}}
 \affiliation{Department of \{Computer Science\sup{1} $|$ Engineering\sup{2}\}, Durham University, Durham, UK}
 }


\abstract{
Anomaly detection is the task of recognising novel samples which deviate significantly from pre-established normality. Abnormal classes are not present during training meaning that models must learn effective representations solely across normal class data samples. Deep Autoencoders (AE) have been widely used for anomaly detection tasks, but suffer from overfitting to a null identity function. To address this problem, we implement a training scheme applied to a Denoising Autoencoder (DAE) which introduces an efficient method of producing Adversarially Learned Continuous Noise (ALCN) to maximally globally corrupt the input prior to denoising. Prior methods have applied similar approaches of adversarial training to increase the robustness of DAE, however they exhibit limitations such as slow inference speed reducing their real-world applicability or producing generalised obfuscation which is more trivial to denoise. We show through rigorous evaluation that our ALCN method of regularisation during training improves AUC performance during inference while remaining efficient over both classical, leave-one-out novelty detection tasks with the variations-: $9$ (normal) vs. $1$ (abnormal) \& $1$ (normal) vs. $9$ (abnormal); MNIST - $AUC_{avg}$: $0.890$ \& $0.989$, CIFAR-10 - $AUC_{avg}$: $0.670$ \& $0.742$, in addition to challenging real-world anomaly detection tasks: industrial inspection (MVTEC-AD - $AUC_{avg}$: $0.780$) and plant disease detection (Plant Village - $AUC$: $0.770$) when compared to prior approaches. 
}

\onecolumn \maketitle \normalsize \setcounter{footnote}{0} \vfill

\section{Introduction} \label{sec:intro}
\noindent The task of anomaly detection is challenging due to deviations from normality being continuous and sporadic by nature. Anomalous space is open-set continuous, meaning that strictly supervised classifiers, although performing well across tasks in anomaly detection \cite{Gaus2019,Bhowmik2019} are restricted by their limited exposure to abnormal examples during training. It is impossible for datasets to contain every possible deviation in the anomalous data thus supervised (classification-based) approaches cannot generalise to the continuous nature in which anomalous samples may deviate from normality. This means that there will always exist anomalous deviations in anomaly space which present as adversarial examples to supervised methods.

Generative-based anomaly detection methods \cite{Schlegl2017,Schlegl2019,zenati2018,Akcay2018,Akcay2019} train solely across normal examples in order to approximate the underlying distribution of normality. They work by learning meaningful features to solely represent normal samples which will cause a relatively small reconstruction error after decoding; conversely, the model will fail to reconstruct anomalous samples fully due to null exposure of the anomalous parts during training. As such, the reconstruction error between input and output provides a sound metric to measure anomalous deviation of presented samples. The benefit of this (semi-supervised) training is that normal (non-anomalous) data is often relatively inexpensive and plentiful to obtain within real-world anomaly detection tasks.

Autoencoders (AE) are well-suited to the approximation of the the underlying data distribution across the normal class. They exhibit stability during training unlike their Generative Adversarial Network (GAN) \cite{Goodfellow2014GAN} based counterparts which exhibit training difficulties such as mode-collapse or convergence instability \cite{GANConvergence_modecollapse}. AE do however risk converging to a pass-through identity function ($\mathbbm{1}$) \cite{bengio2013generalized} for which the mapping from input $x$ to output $x'$ is a null function such that $\lim_{y \rightarrow 0} y=\mathcal{L}(x, x') \Rightarrow  x \simeq x'$ where $\mathcal{L}$ is the reconstruction error. Although this can still learn limited underlying information about the distribution of the training data, this over-fitting allows the reconstruction of anomalous regions within the input which negatively affects performance in the task of semi-supervised anomaly detection. To prevent this, Denoising Autoencoders (DAE) \cite{bengio2013generalized} are trained to produce unperturbed reconstructions from purposefully noised input. This applies a level of regularisation to the AE such that convergence to a trivial solution is not straightforward. It allows an AE to learn more robust and meaningful features across normality as well as remain invariant of noise present in the input \cite{Salehi2021ARAE,Jewell2022OLED}. 

Adding noise to input images in the task of semi-supervised anomaly detection has been explored previously \cite{Salehi2021ARAE,Jewell2022OLED,pathak2016context}. The Adversarially Robust Autoencoder (ARAE) \cite{Salehi2021ARAE} works by forcing perceptually similar samples closer in their latent representations by crafting adversarial examples that are constrained to be 1) perceptually similar to the input, but have 2) maximally distant latent encoding. The adversarial samples are produced by traversing the latent space at each training epoch to find samples which optimally satisfy conditions 1 and 2. This process significantly increases computational overhead of the model due to the demands of satisfying such constraints. As such, the latency of ARAE \cite{Salehi2021ARAE} is slow during training. 

The One-Class Learned Encoder-Decoder (OLED) \cite{Jewell2022OLED} partially obfuscates the input data with a mask produced by an additional autoencoder network called the Mask Module (MM). The MM is optimised to produce masks which maximise the reconstruction error of the DAE module. A limitation of this method is that the produced masks are visually similar across all datasets, becoming, in-essence, a one-size-fits-all type of obfuscation.

In this work, we address the limitations of prior work \cite{Salehi2021ARAE,Jewell2022OLED} by producing tailored noise to the given task efficiently by extending the notion of optimised adversarial noise for robust training with the Adversarially Learned Continuous Noise (ALCN) method. Our method has two parts which are trained simultaneously: 1) The Noise Generator $G_{noise}$ module which produces maximal and continuous noise which is bespoke to the training data and 2) The Denoising Autoencoder $G_{denoise}$ module which is trained to reconstruct input images corrupted (by weighted sum) by the output of $G_{noise}$.

\vspace{0.4cm}
\noindent In this work, we propose the following contributions:

\begin{itemize}
  \item[--] A novel method of adding continuous adversarially generated noise to input images which are optimised to be maximally challenging for a denoising autoencoder to reverse.
  
  \item[--] Exhaustive evaluation of this approach against prior noising methods \cite{Salehi2021ARAE,Jewell2022OLED,pathak2016context} as well as against manually defined noise (Random Speckle and Gaussian) across `leave-one-out' anomaly detection tasks formulated via the MNIST \cite{lecun-mnisthandwrittendigit-2010} and CIFAR-10 \cite{Krizhevsky09learningmultiple} benchmark datasets.
  
  \item[--] Extended evaluation over real-world anomaly detection tasks including industrial inspection (MVTEC \cite{bergmann2019mvtec}) and the plant leaf disease detection (Plant Village \cite{Hughes2015Plant_Village}) with side-by-side comparison against leading state-of-the-art methods \cite{Akcay2018,akccay2019skip,Vu2019ADAE,zenati2018,Schlegl2017,Ruff2018DSVDD,Perera2019OCGAN,Abati2019LSA,Salehi2021ARAE,Jewell2022OLED} via the Area Under Receiver Operator Characteristic (AUC) metric.
  
\end{itemize}


\section{Related Work} \label{sec:relwork}
\noindent Existing anomaly detection methods have gained exceptional success in identifying data instances which deviate significantly from established normality. However, the current methods struggle to address fully the two enduring anomaly detection challenges. Firstly, data availability and coverage is always limited for the anomalous class such that those limited anomaly examples present provide poor coverge of the full sprectrum of possible anomalous deviations. Second, is the challenge of a high-skewed dataset distribution such that normal instances dominate but with anomaly contamination \cite{pang2019deep}. In order to combat these challenges, deep anomaly detection methods operate in a domain of a binary-class, semi-supervised learning paradigm. These are typically trained to solely represent normal class data with varying representations spanning the latent space of Generative Adversarial Networks (GAN) \cite{Schlegl2017,akcay2018ganomaly,zenati2018}, distance metric spaces within  \cite{pang2018learning,ruff2018deep} or intermediate representations via autoencoders \cite{zhou2017anomaly}. Subsequently, these learned representations are used to define normality as an anomaly score correlated to reconstruction error \cite{Schlegl2017,akcay2018ganomaly,zenati2018}  or distance-based measures  \cite{pang2018learning,ruff2018deep}.

Generally, semi-supervised anomaly detection approaches \cite{Schlegl2017,akcay2018ganomaly,akccay2019skip} are based on learning a close approximation to the true distribution of normal instances by using generative methods, such as \cite{akcay2018ganomaly,akccay2019skip,Barker2021}. The initial strategy uses autoencoder \cite{lecun2015deep} architectures such as a variational autoencoder (VAE) \cite{Kingma2013}, where a latent representation $z$ is learned from the image space $X$ via an encoder mapping via $Pr(z|x)$. Sequentially, a decoder maps from $z$ back to image space via $Pr(i'|x)$ to produce $x'$. The encoder and decoder is trained to minimise reconstruction error
between the original image $x \in X$ and the reconstruction image 
$x'$. However, in general, they do not closely capture the data distribution over $X$ due to the oversimplification of the learned prior probability $p(z|x)$. VAE \cite{Kingma2013}
are only capable of learning a uni-modal distribution, which fails to capture complex distributions that are commonplace in real world anomaly detection scenarios \cite{Barker2021}.

AnoGAN \cite{Schlegl2017} combats this simplification by adopting GAN in the anomaly detection approach. AnoGAN \cite{Schlegl2017} is the first GAN-based method, where the model is trained to learn the manifold $z$ only on normal data. When anomalous $x_a$ is going through the generator network ($G$), it produces an $l_2$ reconstruction error which, if large enough from learned normal data distribution will be flagged anomalous. Although effectively proven, the computational performance is prolonged hence limiting real-world applicability. GANomaly \cite{Akcay2018} solves this issue by training an encoder-decoder-encoder network with the adversarial scheme to capture the normal distribution within the image and latent
space. It is achieved by training a generator network and a secondary encoder in
order to map the generated samples into a second latent space $\hat{z}$ which is then used to better learn the original latent priors $z$, mapping between latent values efficiently at the same time as the generator $G$ learns the distribution manifold over data $x$. Efficient GAN Based Anomaly Detection (EGBAD) also addresses the performance issue in AnoGAN by adopting a Bidirectional GAN \cite{Donahue2019} into its architecture. The primary idea is to solve, during training, the optimisation problem $min_{G,E}max_DV(D,G,E)$ where the features of $X$ are learned by the network $E$ to produce the pair of $(x,E(x))$. The main contribution is to allow EGBAD to compute the anomaly score without $\Gamma$ optimisation steps during inference as it happens in AnoGAN \cite{Schlegl2017}.

Although GAN-based methods for anomaly detection have risen to prominence and gained significant results, they suffer from volatile training issues such as mode collapse \cite{TungModecollapse2020}, leading to potential inability for the generator to produce meaningful output. On the other hand, autoencoder \cite{lecun2015deep} based architectures are much more stable than GAN-based approaches, but can overfit to a pass-through identity (null) function as previously discussed. To combat this, regularisation in the form of adding deliberate corruption to the input data often takes place \cite{adey2021autoencoders,Salehi2021ARAE,Jewell2022OLED}.

The work of \cite{adey2021autoencoders} adds purposeful corruption to the normal input data and subsequently forces the autoencoder to reconstruct it, or denoise it. It enables the model to compress anomaly score to zero for normal pixel, resulting clean anomaly segmentation  which  significantly  improve  performance. ARAE \cite{Salehi2021ARAE} works by injecting adversarial samples into the training set so that the model can fit the original sample and the adversarial sample at the same time. It is shown that ARAE \cite{Salehi2021ARAE} learns more semantically meaningful features of normal class by training an adversarially robust autoencoder in a latent space, resulting competitive performance  with  state-of-the-art in novelty detection.  

The work of OLED \cite{Jewell2022OLED} offers another approach in noise perturbation in input data, where instead of being perturbed by noise, input images are subjected to masking through the use of Mask Module (MM). The masks generated by MM are optimized to cover the most important parts of the input image,  resulting in a comparable reconstruction score across sample. Through optimal masking, the proposed approach learns semantically richer representations and enhances novelty detection at test time. 

Motivated by the idea intention pre-encoding input corruption, we propose a novel approach for adversarially generated noise which, when added to the input data, is very challenging for the denoising autoencoder to reverse. Our approach, Adversarially Learned Continuous Noise (ALCN), it consists of two parts, Noise Generator $G_{noise}$ and Denoising Autoencoder $G_{denoise}$. The former produces maximal and continuous noise which is bespoke to the training data while the latter trained to reconstruct input images perturbed (by weighted sum) with this maximal noise.




\section{Proposed Approach}  \label{sec:proposal}

 \begin{figure*}[!htb]
\centering
\includegraphics[width=\linewidth]{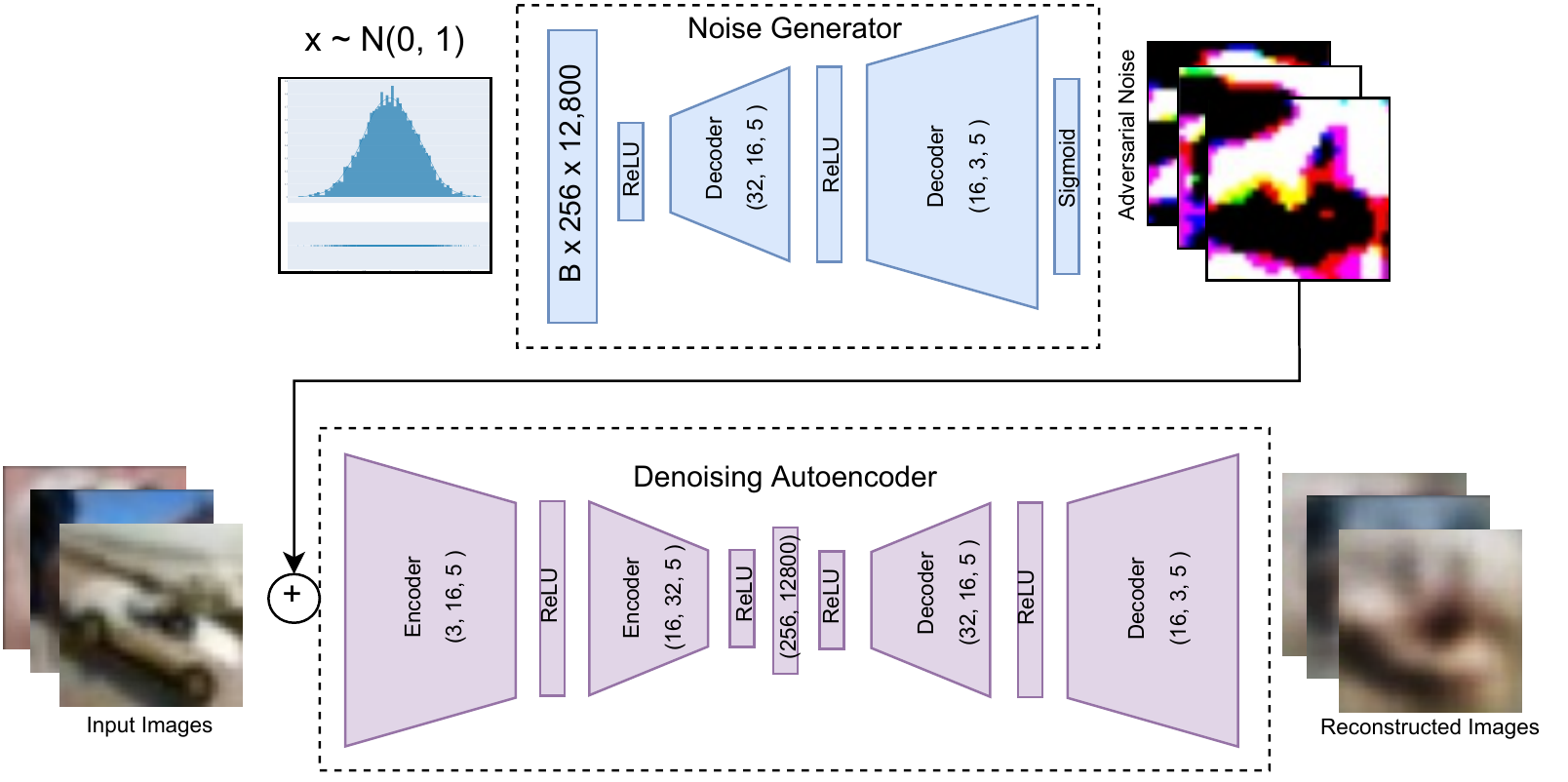}
\caption{Overview of adversarial noise learning architecture featuring: top-Noise Generator Module $G_{noise}$, bottom- Denoising module $G_{denoise}$.}
\label{fig:model}
\end{figure*}

Our proposed method is outlined in Figure \ref{fig:model}. In our approach, we utilise a Denoising Autoencoder Generator ($G_{denoise}$) network together with a GAN-like Noise Generator ($G_{noise}$) network. These are concurrently adversarially trained using the process outlined in Algorithm \ref{alg:cap}. In a given step, the weights of $G_{noise}$ are updated first with gradient ascent with respect to the reconstruction error so that at any given step, $G_{noise}$ produces differing corruption from the previous step which $G_{denoise}$ then attempts to reverse by optimising the reconstruction error with gradient descent.

Training across dataset $x \in \mathbb{R}^{B \times C \times H \times W} \in X$ where $\{B, C, H, W\}$ represent the batch size, number of channels, height and width, respectively, starts by training the noise generator $G_{noise}$. A linear vector of size $B \times 256$ random variables $\phi$ is sampled from a standard Gaussian normal distribution $\phi \sim N(\mu:0, \sigma:1)$ and fed through $G_{noise}$ to produce noise $n$ of shape 
$\mathbb{R}^{B \times C \times H \times W}$. The added Sigmoid layer 
($\frac{1}{1+e^{-l}}$) on the final layer of $G_{noise}$ binds the noise values continuously between [$0, 1$]. We combine the noise $n$ to the input image $x$ using a weighted sum by utilising the linear blending operator $noise(x, n) = \alpha (x) + (1-\alpha)(n)$ where $\alpha$ is randomly sampled on each step within bounds $\alpha \rightarrow [0.2, 0.9] \in \mathbb{R}^{+}$. The linear blend operator ensures that the magnitude of the values of $noise(x,n)$ match with the pixel intensities of $x$ and $n$. Values of $x$ are normalised with $0$ mean and unit variance meaning that the values of $noise(x,n)$ are such that $G_{denoise}$ is prevented from discriminating between the noise corrupted pixels and the original image pixels based on differing pixel intensity.

\begin{algorithm}[ht!]
\caption{Adversarial Noise Training}\label{alg:cap}
\begin{algorithmic}

\State $\text{W}\{G \} \gets \text{init}$ \Comment{Initialise G randomly}

\State $\text{W}\{N_{G} \} \gets \text{init}$ \Comment{Initialise $N_{G}$ randomly}

\flushleft \emph{Train One Epoch}:

\For {mini-batch: $x \subset X$}

weights$\{G_{noise} \} \gets$ True

weights$\{G_{denoise} \} \gets$ False

$\alpha \leftarrow [0.2, 0.9]$ \Comment{Randomly select $\alpha$}

$\text{z} \gets N(\mu = 0, \sigma = 1)$  \Comment{$|z| = \{|x|, 256\}$}

$x' \gets G_{denoise}((1-\alpha) G_{noise}(z) + \alpha x)$

$\text{W} \{G_{noise}\}\xleftarrow{\text{backpropagate}}\text{Optim}_{G_{noise}}(-\mathcal{L}(x, x'))$ 

weights$\{G_{noise} \} \gets$ False

weights$\{G_{denoise} \} \gets$ True

$x' \gets G_{denoise}(G_{noise}(z) + x)$

$\text{W} \{ G_{denoise} \} \xleftarrow{\text{backpropagate}} \text{Optim}_{G_{denoise}}(\mathcal{L}(x, x'))$ 

\EndFor{}
\end{algorithmic}
\end{algorithm}

 If alpha is static during training, $G_{noise}$ can theoretically perfectly optimise the generated noise $n$ to destroy all information in image $x \in X$ such that all values in $noise(x,n)$ are set to 1 such that $n = (\frac{1-\alpha \cdot x \in X} {1-\alpha})$. The $noise(x,n)$ cannot converge to all zeros where $n = -(\frac{\alpha \cdot x \in X}{1-\alpha})$ due to the logical argument that the values of noise $n$ produced by $G_{noise}$ are bound to [0,1] $\in \mathbb{R}^{+}$ because of the Sigmoid layer on the output of $G_{noise}$ and $x$ is such that $\forall x_{i} \in x \rightarrow \{0,1\}, \exists x_{i} \in x \, | \, x_{i}=1$ implying that if $(x_{i} \in x = 1)$ then $ n = \frac{-\alpha}{1-\alpha} \Rightarrow  n < 0  \,\, \forall \alpha \therefore n \notin \mathbb{R}^{+}$. To prevent convergence to the trivial solution $n = (\frac{1-\alpha \cdot x \in X} {1-\alpha})$ in our experiments, we: 1) Set the value of $\alpha$ to be randomly continuously sampled for each step during training and 2) The input of $G_{noise}$ is sampled from the Gaussian distribution $N(0,1)$ which applies some level of randomness during sampling.

The $noise(x,n)$ is then used as input to $G_{denoise}$ to reconstruct $x$ from $noise(x,n)$, reversing the corruption caused by $G_{noise}$. The corrupted image $noise_{x,n}$ is encoded to the latent vector $z$ and then subsequently decoded into a synthetic reconstruction $x'$.

Adversarial learning is accomplished by the mini-max optimisation between the $G_{denoise}$ and $G_{noise}$ modules. Weights of $G_{denoise}$ are optimised to minimise $\mathcal{L}$, the reconstruction error between $x$ and $x'$ whereas the weights of $G_{noise}$ are conversely optimised to maximise $\mathcal{L}$. Loss terms in the overall loss are given scalar regularisation terms $\lambda_{0}$ and $\lambda_{1}$ for losses $\mathcal{L}_{G_{denoise}}$ and $\mathcal{L}_{G_{noise}}$ respectively. The overall optimisation function in this work is: 

\begin{equation}
     \begin{matrix}argmin & argmax\\ G_{denoise} & G_{noise}\end{matrix} = \mathcal{L}_{G_{denoise}}(x, x')\lambda_{0} + \mathcal{L}_{G_{noise}}(x, x')\lambda_{1} 
\end{equation}

This method of training encourages the noise generator to produce masks which optimally corrupt the input. Such optimal noise makes the denoising process more difficult as the denoising module must not only learn meaningful features of the input data, but such learned representations should not carry forward out-of-distribution (anomalous) features to the synthetic reconstruction.


\subsection{Loss Function}
In our experiments, we find that the use of Focal Frequency Loss (FFL) \cite{jiang2021focal} created higher-fidelity reconstructions and a slight increase in AUC performance. FFL is based on the L2 distance (loss) between the real image $x$ and the generated image $x'$ in the Fourier (frequency) domain. Pixel coordinates of $x$ ($x_{i}$) and $x'$ ($x'_{i}$) are used in conjunction to their respective frequency spectrum coordinates $(x_{i}^{freq}$ \& $x'$$_{i}^{freq})$ from the Discrete Fourier Transform (DFT) as follows:

\begin{equation}
F(\vartheta )  = (\frac{\vartheta_{i} \cdot \vartheta _{i}^{freq}} {|H|}) | \vartheta  = \{x, x'\}
\end{equation}

The loss is defined as the total distance in frequency domain with respect to amplitude and phase in the following formula: 

\begin{equation}
\mathcal{L}(x, x') = ||e^{-i2\pi(F(x))} - e^{-i2\pi(F(x'))}||^2
\end{equation}

\begin{figure}[htb!]
 \centering
\includegraphics[width=200px]{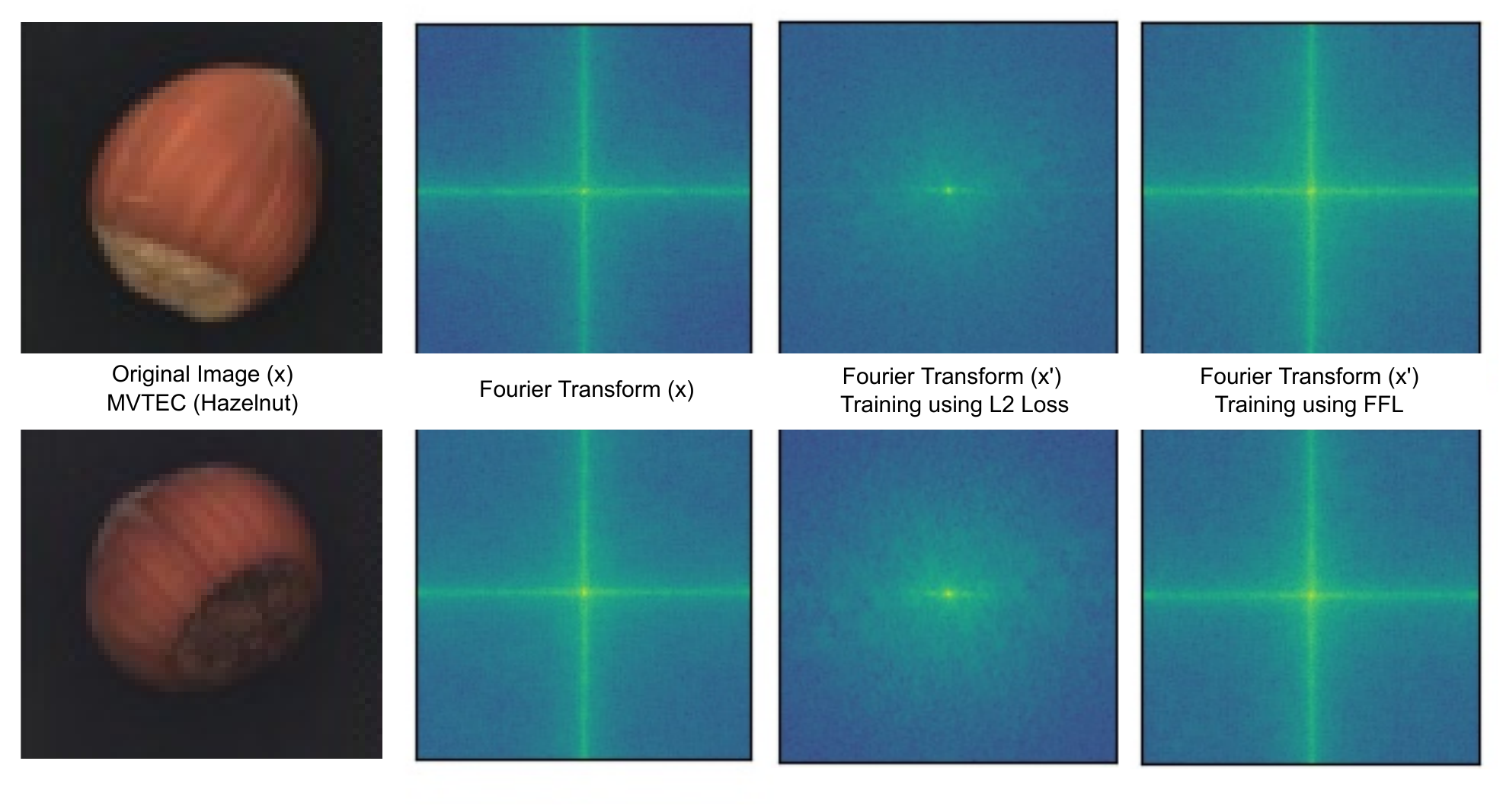}
\caption{Visualisation of frequency domain after Fourier Transform operation of reconstruction $x'$ from input image $x$ both with and without using FFL \cite{jiang2021focal} during training. }
\label{fig:fourier}
\end{figure}

Figure \ref{fig:fourier} shows visually how using an L2 loss loosely approximates the frequency representation of $x$, but fails to capture high-frequency information present in the image. FFL \cite{jiang2021focal} however, can more closely approximate the frequency domain as seen in this figure, the frequency representations of $x$ and $x'$ are closely matched. This property makes it highly suitable for use in our reconstruction-driven anomaly detection approach.

\section{Experimental Setup} \label{sec:eva}


\begin{table*}[htb!]

\centering
\caption{Quantitative results (class name indicates AUC, $AUC_{avg}$ of all classes) of models across MNIST \cite{lecun-mnisthandwrittendigit-2010} (upper) and CIFAR-10 \cite{Krizhevsky09learningmultiple} (lower) datasets (Protocol 1).}
\resizebox{\linewidth}{!}{
\begin{tabular}{|l|lllllllllll|}

\hline
\multirow{2}{*}{{\bf Model}} &
  \multicolumn{11}{c|}{\textbf{MNIST}} \\ \cline{2-12}
 &
  \multicolumn{1}{l|}{\textbf{0}} &
  \multicolumn{1}{l|}{\textbf{1}} &
  \multicolumn{1}{l|}{\textbf{2}} &
  \multicolumn{1}{l|}{\textbf{3}} &
  \multicolumn{1}{l|}{\textbf{4}} &
  \multicolumn{1}{l|}{\textbf{5}} &
  \multicolumn{1}{l|}{\textbf{6}} &
  \multicolumn{1}{l|}{\textbf{7}} &
  \multicolumn{1}{l|}{\textbf{8}} &
  \multicolumn{1}{l|}{\textbf{9}} &
  \textbf{$AUC_{avg}$} \\ \hline \hline
  VAE \cite{Kingma2013} & 
  \multicolumn{1}{l|}{0.55} &
  \multicolumn{1}{l|}{0.10} &
  \multicolumn{1}{l|}{0.63} &
  \multicolumn{1}{l|}{0.25} &
  \multicolumn{1}{l|}{0.35} &
  \multicolumn{1}{l|}{0.30} &
  \multicolumn{1}{l|}{0.43} &
  \multicolumn{1}{l|}{0.18} &
  \multicolumn{1}{l|}{0.50} &
  \multicolumn{1}{l|}{0.10} &
  0.34 \\ \hline
AnoGAN \cite{Schlegl2017} &
  \multicolumn{1}{l|}{0.61} &
  \multicolumn{1}{l|}{0.30} &
  \multicolumn{1}{l|}{0.54} &
  \multicolumn{1}{l|}{0.44} &
  \multicolumn{1}{l|}{0.43} &
  \multicolumn{1}{l|}{0.42} &
  \multicolumn{1}{l|}{0.48} &
  \multicolumn{1}{l|}{0.36} &
  \multicolumn{1}{l|}{0.40} &
  \multicolumn{1}{l|}{0.34} &
  0.43 \\ \hline
EGBAD \cite{zenati2018} &
  \multicolumn{1}{l|}{0.78} &
  \multicolumn{1}{l|}{0.29} &
  \multicolumn{1}{l|}{0.67} &
  \multicolumn{1}{l|}{0.52} &
  \multicolumn{1}{l|}{0.45} &
  \multicolumn{1}{l|}{0.43} &
  \multicolumn{1}{l|}{0.57} &
  \multicolumn{1}{l|}{0.40} &
  \multicolumn{1}{l|}{0.55} &
  \multicolumn{1}{l|}{0.35} &
  0.50 \\ \hline
GANomaly \cite{Akcay2018} &
  \multicolumn{1}{l|}{0.89} &
  \multicolumn{1}{l|}{0.65} &
  \multicolumn{1}{l|}{0.93} &
  \multicolumn{1}{l|}{0.80} &
  \multicolumn{1}{l|}{0.82} &
  \multicolumn{1}{l|}{0.85} &
  \multicolumn{1}{l|}{0.84} &
  \multicolumn{1}{l|}{0.69} &
  \multicolumn{1}{l|}{0.87} &
  \multicolumn{1}{l|}{0.55} &
  0.79 \\ \hline
  
ADAE \cite{Vu2019} &
  \multicolumn{1}{l|}{0.95} &
  \multicolumn{1}{l|}{0.82} &
  \multicolumn{1}{l|}{0.95} &
  \multicolumn{1}{l|}{0.89} &
  \multicolumn{1}{l|}{0.82} &
  \multicolumn{1}{l|}{\textbf{0.91}} &
  \multicolumn{1}{l|}{0.89} &
  \multicolumn{1}{l|}{0.80} &
  \multicolumn{1}{l|}{0.93} &
  \multicolumn{1}{l|}{0.63} &
  0.86 \\ \hline


DAE &
  \multicolumn{1}{l|}{0.84} &
  \multicolumn{1}{l|}{0.97} &
  \multicolumn{1}{l|}{0.79} &
  \multicolumn{1}{l|}{0.64} &
  \multicolumn{1}{l|}{0.53} &
  \multicolumn{1}{l|}{0.61} &
  \multicolumn{1}{l|}{0.66} &
  \multicolumn{1}{l|}{0.55} &
  \multicolumn{1}{l|}{0.71} &
  \multicolumn{1}{l|}{0.57} &
  0.69 \\ \hline

DAE+Random Noise &
  \multicolumn{1}{l|}{0.84} &
  \multicolumn{1}{l|}{0.93} &
  \multicolumn{1}{l|}{0.66} &
  \multicolumn{1}{l|}{0.66} &
  \multicolumn{1}{l|}{0.52} &
  \multicolumn{1}{l|}{0.62} &
  \multicolumn{1}{l|}{0.72} &
  \multicolumn{1}{l|}{0.56} &
  \multicolumn{1}{l|}{0.75} &
  \multicolumn{1}{l|}{0.53} &
  0.68 \\ \hline
\begin{tabular}[c]{@{}l@{}}DAE+Gaussian Noise\\ $\sim N(0, 0.5)$ \end{tabular} &
  \multicolumn{1}{l|}{0.88} &
  \multicolumn{1}{l|}{0.97} &
  \multicolumn{1}{l|}{0.77} &
  \multicolumn{1}{l|}{0.66} &
  \multicolumn{1}{l|}{0.55} &
  \multicolumn{1}{l|}{0.62} &
  \multicolumn{1}{l|}{0.75} &
  \multicolumn{1}{l|}{0.55} &
  \multicolumn{1}{l|}{0.71} &
  \multicolumn{1}{l|}{0.57} &
  0.70 \\ \hline
\textbf{DAE + ALCN} &
  \multicolumn{1}{l|}{\textbf{0.97}} &
  \multicolumn{1}{l|}{\textbf{0.97}} &
  \multicolumn{1}{l|}{\textbf{0.96}} &
  \multicolumn{1}{l|}{\textbf{0.89}} &
  \multicolumn{1}{l|}{\textbf{0.85}} &
  \multicolumn{1}{l|}{{0.88}} &
  \multicolumn{1}{l|}{\textbf{0.92}} &
  \multicolumn{1}{l|}{\textbf{0.80}} &
  \multicolumn{1}{l|}{\textbf{0.93}} &
  \multicolumn{1}{l|}{\textbf{0.76}} &
  \textbf{0.89} \\ \hline \hline 
 
 \multirow{2}{*}{{\bf Model}} & \multicolumn{11}{c|}{\textbf{CIFAR-10}} \\ \cline{2-12}

& \multicolumn{1}{l|}{\textbf{Plane}} &
  \multicolumn{1}{l|}{\textbf{Car}} &
  \multicolumn{1}{l|}{\textbf{Bird}} &
  \multicolumn{1}{l|}{\textbf{Cat}} &
  \multicolumn{1}{l|}{\textbf{Deer}} &
  \multicolumn{1}{l|}{\textbf{Dog}} &
  \multicolumn{1}{l|}{\textbf{Frog}} &
  \multicolumn{1}{l|}{\textbf{Horse}} &
  \multicolumn{1}{l|}{\textbf{Ship}} &
  \multicolumn{1}{l|}{\textbf{Truck}} &
  \textbf{$AUC_{avg}$} \\ \hline \hline
VAE \cite{Kingma2013} &
  \multicolumn{1}{l|}{0.59} &
  \multicolumn{1}{l|}{0.40} &
  \multicolumn{1}{l|}{0.52} &
  \multicolumn{1}{l|}{0.44} &
  \multicolumn{1}{l|}{0.46} &
  \multicolumn{1}{l|}{0.50} &
  \multicolumn{1}{l|}{0.38} &
  \multicolumn{1}{l|}{0.51} &
  \multicolumn{1}{l|}{0.64} &
  \multicolumn{1}{l|}{0.49} &
  0.49 \\ \hline
AnoGAN \cite{Schlegl2017} &
  \multicolumn{1}{l|}{0.51} &
  \multicolumn{1}{l|}{0.49} &
  \multicolumn{1}{l|}{0.41} &
  \multicolumn{1}{l|}{0.40} &
  \multicolumn{1}{l|}{0.34} &
  \multicolumn{1}{l|}{0.39} &
  \multicolumn{1}{l|}{0.34} &
  \multicolumn{1}{l|}{0.41} &
  \multicolumn{1}{l|}{0.56} &
  \multicolumn{1}{l|}{0.51} &
  0.44 \\ \hline
EGBAD \cite{zenati2018} &
  \multicolumn{1}{l|}{0.58} &
  \multicolumn{1}{l|}{0.52} &
  \multicolumn{1}{l|}{0.39} &
  \multicolumn{1}{l|}{0.45} &
  \multicolumn{1}{l|}{0.37} &
  \multicolumn{1}{l|}{0.49} &
  \multicolumn{1}{l|}{0.36} &
  \multicolumn{1}{l|}{0.54} &
  \multicolumn{1}{l|}{0.42} &
  \multicolumn{1}{l|}{0.55} &
  0.47 \\ \hline
GANomaly \cite{Akcay2018} &
  \multicolumn{1}{l|}{0.63} &
  \multicolumn{1}{l|}{0.63} &
  \multicolumn{1}{l|}{0.51} &
  \multicolumn{1}{l|}{\textbf{0.58}} &
  \multicolumn{1}{l|}{0.59} &
  \multicolumn{1}{l|}{0.62} &
  \multicolumn{1}{l|}{0.68} &
  \multicolumn{1}{l|}{\textbf{0.61}} &
  \multicolumn{1}{l|}{0.62} &
  \multicolumn{1}{l|}{0.62} &
  0.61 \\ \hline
  
ADAE \cite{Vu2019ADAE} &
  \multicolumn{1}{l|}{0.63} &
  \multicolumn{1}{l|}{\textbf{0.73}} &
  \multicolumn{1}{l|}{0.55} &
  \multicolumn{1}{l|}{\textbf{0.58}} &
  \multicolumn{1}{l|}{0.50} &
  \multicolumn{1}{l|}{0.60} &
  \multicolumn{1}{l|}{0.60} &
  \multicolumn{1}{l|}{\textbf{0.61}} &
  \multicolumn{1}{l|}{0.62} &
  \multicolumn{1}{l|}{0.67} &
  0.61 \\ \hline


DAE &
  \multicolumn{1}{l|}{0.50} &
  \multicolumn{1}{l|}{0.68} &
  \multicolumn{1}{l|}{0.61} &
  \multicolumn{1}{l|}{0.55} &
  \multicolumn{1}{l|}{0.69} &
  \multicolumn{1}{l|}{0.53} &
  \multicolumn{1}{l|}{0.62} &
  \multicolumn{1}{l|}{0.60} &
  \multicolumn{1}{l|}{0.63} &
  \multicolumn{1}{l|}{\textbf{0.71}} &
  0.61 \\ \hline

DAE+Random Noise &
  \multicolumn{1}{l|}{0.63} &
  \multicolumn{1}{l|}{0.53} &
  \multicolumn{1}{l|}{0.54} &
  \multicolumn{1}{l|}{0.54} &
  \multicolumn{1}{l|}{0.65} &
  \multicolumn{1}{l|}{0.59} &
  \multicolumn{1}{l|}{0.64} &
  \multicolumn{1}{l|}{0.55} &
  \multicolumn{1}{l|}{0.66} &
  \multicolumn{1}{l|}{0.63} &
  0.60 \\ \hline
\begin{tabular}[c]{@{}l@{}}DAE+Gaussian Noise\\ $\sim N(0, 0.5)$\end{tabular} &
  \multicolumn{1}{l|}{0.57} &
  \multicolumn{1}{l|}{0.68} &
  \multicolumn{1}{l|}{0.57} &
  \multicolumn{1}{l|}{0.54} &
  \multicolumn{1}{l|}{0.65} &
  \multicolumn{1}{l|}{0.54} &
  \multicolumn{1}{l|}{0.55} &
  \multicolumn{1}{l|}{0.52} &
  \multicolumn{1}{l|}{0.57} &
  \multicolumn{1}{l|}{0.53} &
  0.57 \\ \hline
\textbf{DAE + ALCN} &
  \multicolumn{1}{l|}{\textbf{0.77}} &
  \multicolumn{1}{l|}{0.71} &
  \multicolumn{1}{l|}{\textbf{0.62}} &
  \multicolumn{1}{l|}{0.57} &
  \multicolumn{1}{l|}{\textbf{0.72}} &
  \multicolumn{1}{l|}{\textbf{0.62}} &
  \multicolumn{1}{l|}{\textbf{0.72}} &
  \multicolumn{1}{l|}{0.60} &
  \multicolumn{1}{l|}{\textbf{0.66}} &
  \multicolumn{1}{l|}{0.69} &
  \textbf{0.67} \\ \hline
\end{tabular}
}
\label{tab:MNISTCIFAR}
\end{table*}

\noindent We present our experimental setup in terms of the benchmark datasets used for evaluation (Section \ref{dataset}) and the implementation details of our approach (Section \ref{implementation}).

\subsection{Datasets} \label{dataset}

\noindent We make use of four established benchmark datasets that are commonplace for evaluation within the anomaly detection domain:

\begin{itemize}
    
\item  \textbf{MNIST \cite{lecun-mnisthandwrittendigit-2010}} : A collection of $69,018$ hand-written single digits from $0$ to $9$ of resolution $28 \times 28$. For this dataset we utilise a $80:20$ ($55,209:13,807$) split between training and testing respectively across the data. 

\item  \textbf{CIFAR-10 \cite{Krizhevsky09learningmultiple}} : A set of $50,026$ low-resolution ($32 \times 32$) images split into ten classes of common objects. A $80:20$ ($40,012:10,012$) split between training and testing sets are utilised across this dataset.

\item  \textbf{MVTEC-AD \cite{bergmann2019mvtec}} : Benchmark dataset of $6,809$ images for quality control in industrial visual inspection. The data is composed of fifteen classes of both non-anomalous, defect free objects as well as a set of defective anomalous counter-parts. A $70:30$ split for training and testing respectively is applied for each class. 

\item{\textbf{Plant Village \cite{Hughes2015Plant_Village}}}: Visual images of the leaves of vital agricultural edible plants together with anomalies containing common visual leaf diseases for each respective plant.  

\end{itemize}

\subsection{Implementation Details} \label{implementation}
\noindent Our method is compared across the MNIST \cite{lecun-mnisthandwrittendigit-2010} and CIFAR-10 \cite{Krizhevsky09learningmultiple} datasets due to their inherent simplicity while training as well as giving sufficient bench-marking for the evaluation between the techniques included in this work. Evaluation is conducted in two protocols following from established methods for `leave-one-out' anomaly detection tasks. During protocol 1 (1 vs. rest), one digit is regarded as anomalous and remaining classes are normal as performed by: \cite{akcay2018ganomaly,akccay2019skip,Barker2021,zenati2018,Schlegl2017,Schlegl2019}.
Protocol 2 (rest vs. 1) as performed by: \cite{Ruff2018DSVDD,Perera2019OCGAN,Abati2019LSA,Salehi2021ARAE,Jewell2022OLED} is the opposite in that one digit is normal and the nine remaining classes are anomalous. 

The split ratio for the data is $80:20$ for training and testing respectively as conducted by \cite{zenati2018,Akcay2018}. During training, the Adam optimiser is used for both $G_{denoise}$ and $G_{noise}$ with learning rates of $1 \times 10^{-5}$ and $8 \times 10^{-3}$ respectively. An image resolution of $28\times28$ is implemented throughout `leave-one-out' anomaly detection tasks \cite{lecun-mnisthandwrittendigit-2010,Krizhevsky09learningmultiple}. We implement a larger resolution of $256 \times 256$ across MVTEC \cite{bergmann2019mvtec} and Plant Village \cite{Hughes2015Plant_Village} however. A batch size of 4096 is employed across MNIST and CIFAR-10 and a batch size of 16 is used across MVTEC and plant village during training on an NVidia GTX 1080 TI GPU. We evaluate our method using the Area Under Receiver Operator Characteristic (AUC) metric.

\begin{table}[htb!]
\centering
\caption{Quantitative results ($AUC_{avg}$) of models including ARAE \cite{Salehi2021ARAE} and OLED \cite{Jewell2022OLED} across MNIST \cite{lecun-mnisthandwrittendigit-2010} (left) and CIFAR-10 \cite{Krizhevsky09learningmultiple} (right) datasets (Protocol 2).}
\addtolength{\tabcolsep}{-4.0pt}
\begin{tabular}{l|c|c|}
\cline{2-3}
\multicolumn{1}{c|}{}        & \textbf{MNIST}                              & \textbf{CIFAR-10}       \\ \hline
\multicolumn{1}{|l|}{\textbf{Method}} & \multicolumn{1}{l|}{\textbf{AUC$_{avg}$}} & {\textbf{AUC$_{avg}$}} \\ \hline \hline
\multicolumn{1}{|l|}{DSVDD \cite{Ruff2018DSVDD}}                 & 0.948                                       & 0.648                   \\ \hline
\multicolumn{1}{|l|}{OCGAN \cite{Perera2019OCGAN}}                 & 0.975                                       & 0.733                   \\ \hline
\multicolumn{1}{|l|}{LSA\cite{Abati2019LSA}}                   & 0.975                                       & 0.731                   \\ \hline
\multicolumn{1}{|l|}{ARAE \cite{Salehi2021ARAE}}                  & 0.975                                       & 0.717                   \\ \hline
\multicolumn{1}{|l|}{OLED \cite{Jewell2022OLED}}                  & 0.985                                       & 0.671                   \\ \hline
\multicolumn{1}{|l|}{\textbf{DAE + ALCN}}          & \textbf{0.989}                              & \textbf{0.742}          \\ \hline

\end{tabular}
\addtolength{\tabcolsep}{0pt}  

\label{tab:protocol2MNIST-CIFAR}
\end{table}
\section{Results} \label{sec:results}

\begin{table*}[htb!]

\centering
\caption{Quantitative results (class name indicates AUC, $AUC_{avg}$ of all classes) of models across MVTEC-AD \cite{bergmann2019mvtec} dataset.}

\resizebox{\linewidth}{!}{
\begin{tabular}{|l|llllllllllllllll|}
\hline
\multirow{2}{*}{\textbf{Model}} &
  \multicolumn{16}{c|}{\textbf{MVTEC-AD}} \\ \cline{2-17} 
 &
  \multicolumn{1}{c|}{\textbf{Bottle}} &
  \multicolumn{1}{c|}{\textbf{Cable}} &
  \multicolumn{1}{c|}{\textbf{Caps.}} &
  \multicolumn{1}{c|}{\textbf{Carpet}} &
  \multicolumn{1}{c|}{\textbf{Grid}} &
  \multicolumn{1}{c|}{\textbf{H'nut}} &
  \multicolumn{1}{c|}{\textbf{Leath.}} &
  \multicolumn{1}{c|}{\textbf{M'nut}} &
  \multicolumn{1}{c|}{\textbf{Pill}} &
  \multicolumn{1}{c|}{\textbf{Screw}} &
  \multicolumn{1}{c|}{\textbf{Tile}} &
  \multicolumn{1}{c|}{\textbf{T'brush}} &
  \multicolumn{1}{c|}{\textbf{T'sistor}} &
  \multicolumn{1}{c|}{\textbf{Wood}} &
  \multicolumn{1}{c|}{\textbf{Zipper}} &
  \multicolumn{1}{c|}{\textbf{$AUC_{avg}$}} \\ \hline \hline
VAE \cite{Kingma2013} &
  \multicolumn{1}{l|}{0.66} &
  \multicolumn{1}{l|}{0.63} &
  \multicolumn{1}{l|}{0.61} &
  \multicolumn{1}{l|}{0.51} &
  \multicolumn{1}{l|}{0.52} &
  \multicolumn{1}{l|}{0.30} &
  \multicolumn{1}{l|}{0.41} &
  \multicolumn{1}{l|}{0.66} &
  \multicolumn{1}{l|}{0.51} &
  \multicolumn{1}{l|}{1} &
  \multicolumn{1}{l|}{0.21} &
  \multicolumn{1}{l|}{0.30} &
  \multicolumn{1}{l|}{0.65} &
  \multicolumn{1}{l|}{0.87} &
  \multicolumn{1}{l|}{\textbf{0.87}} &
  0.58 \\ \hline
AnoGAN \cite{Schlegl2017} &
  \multicolumn{1}{l|}{0.80} &
  \multicolumn{1}{l|}{0.48} &
  \multicolumn{1}{l|}{0.44} &
  \multicolumn{1}{l|}{0.34} &
  \multicolumn{1}{l|}{0.87} &
  \multicolumn{1}{l|}{0.26} &
  \multicolumn{1}{l|}{0.45} &
  \multicolumn{1}{l|}{0.28} &
  \multicolumn{1}{l|}{0.71} &
  \multicolumn{1}{l|}{1} &
  \multicolumn{1}{l|}{0.40} &
  \multicolumn{1}{l|}{0.44} &
  \multicolumn{1}{l|}{0.69} &
  \multicolumn{1}{l|}{0.57} &
  \multicolumn{1}{l|}{0.72} &
  0.56 \\ \hline
EGBAD \cite{zenati2018} &
  \multicolumn{1}{l|}{0.63} &
  \multicolumn{1}{l|}{0.68} &
  \multicolumn{1}{l|}{0.52} &
  \multicolumn{1}{l|}{0.52} &
  \multicolumn{1}{l|}{0.54} &
  \multicolumn{1}{l|}{0.43} &
  \multicolumn{1}{l|}{0.55} &
  \multicolumn{1}{l|}{0.47} &
  \multicolumn{1}{l|}{0.57} &
  \multicolumn{1}{l|}{0.43} &
  \multicolumn{1}{l|}{0.79} &
  \multicolumn{1}{l|}{0.64} &
  \multicolumn{1}{l|}{0.73} &
  \multicolumn{1}{l|}{0.91} &
  \multicolumn{1}{l|}{0.58} &
  0.60 \\ \hline
GANomaly \cite{Akcay2018} &
  \multicolumn{1}{l|}{0.89} &
  \multicolumn{1}{l|}{0.76} &
  \multicolumn{1}{l|}{0.73} &
  \multicolumn{1}{l|}{0.70} &
  \multicolumn{1}{l|}{0.71} &
  \multicolumn{1}{l|}{{0.79}} &
  \multicolumn{1}{l|}{{0.84}} &
  \multicolumn{1}{l|}{0.70} &
  \multicolumn{1}{l|}{0.74} &
  \multicolumn{1}{l|}{0.75} &
  \multicolumn{1}{l|}{0.79} &
  \multicolumn{1}{l|}{0.65} &
  \multicolumn{1}{l|}{{0.79}} &
  \multicolumn{1}{l|}{0.83} &
  \multicolumn{1}{l|}{0.75} &
  0.76 \\ \hline

Skip-GANomaly \cite{akccay2019skip} &
  \multicolumn{1}{l|}{0.93} &
  \multicolumn{1}{l|}{0.67} &
  \multicolumn{1}{l|}{0.71} &
  \multicolumn{1}{l|}{0.79} &
  \multicolumn{1}{l|}{0.65} &
  \multicolumn{1}{l|}{0.90} &
  \multicolumn{1}{l|}{\textbf{0.90}} &
  \multicolumn{1}{l|}{0.79} &
  \multicolumn{1}{l|}{0.75} &
  \multicolumn{1}{l|}{1} &
  \multicolumn{1}{l|}{\bf 0.85} &
  \multicolumn{1}{l|}{\bf 0.68} &
  \multicolumn{1}{l|}{\textbf{0.81}} &
  \multicolumn{1}{l|}{0.91} &
  \multicolumn{1}{l|}{0.66} &
  {0.80} \\ \hline
  
\begin{tabular}[c]{@{}l@{}}\textbf{DAE+ALCN}\end{tabular} &
  \multicolumn{1}{l|}{\textbf{0.94}} &
  \multicolumn{1}{l|}{\textbf{0.84}} &
  \multicolumn{1}{l|}{\textbf{0.86}} &
  \multicolumn{1}{l|}{\textbf{0.84}} &
  \multicolumn{1}{l|}{\textbf{0.97}} &
  \multicolumn{1}{l|}{\textbf{0.92}} &
  \multicolumn{1}{l|}{0.62} &
  \multicolumn{1}{l|}{\textbf{0.86}} &
  \multicolumn{1}{l|}{\textbf{0.75}} &
  \multicolumn{1}{l|}{\textbf{1}} &
  \multicolumn{1}{l|}{{0.79}} &
  \multicolumn{1}{l|}{{0.65}} &
  \multicolumn{1}{l|}{0.73} &
  \multicolumn{1}{l|}{\textbf{0.93}} &
  \multicolumn{1}{l|}{0.70} &
  \textbf{0.83} \\ \hline
\end{tabular}
}
\vspace{-0.1cm}
\label{tab:MVTEC}

\end{table*}

\noindent Extensive comparison of the results of our method compared to prior methods are outlined in Tables \ref{tab:MNISTCIFAR}, \ref{tab:protocol2MNIST-CIFAR}, \ref{tab:MVTEC}, \ref{tab:plantvillage} and \ref{tab:complexity}. Tables \ref{tab:MNISTCIFAR} and \ref{tab:protocol2MNIST-CIFAR} outline the quantitative results of the DAE+ALCN method across both MNIST \cite{lecun-mnisthandwrittendigit-2010} and CIFAR-10 \cite{Krizhevsky09learningmultiple} `leave-one-out tasks' across both protocol 1 (9 normal/1 anomalous) and protocol 2 (1 normal/9 anomalous). Across the real-world anomaly detection tasks outlined in this paper \cite{bergmann2019mvtec,Hughes2015Plant_Village}, Table \ref{tab:MVTEC} outlines the quantitative results of our method across the MVTEC-AD \cite{bergmann2019mvtec} industrial inspection dataset and Table \ref{tab:plantvillage} presents the results across the Plant Village dataset \cite{Hughes2015Plant_Village}.

\subsection{Leave One Out Anomaly Detection} \label{subsec:LOOAD}
\subsubsection{Protocol 1}
\noindent Table \ref{tab:MNISTCIFAR} outlines the results of each approach across the MNIST and CIFAR-10 datasets. We begin by comparing our vanilla DAE approach without any noise regularisation and this results in an AUC$_{avg}$ of $0.69$ across MNIST and $0.61$ across CIFAR-10. This is weak compared to other methods in the table. Applying Gaussian noise obtains an AUC$_{avg}$ of $0.70$ on MNIST and $0.57$ on CIFAR-10. Our DAE+ALCN approach applied to the DAE architecture achieves the best AUC score on 90\% of the classes with an average AUC of $0.89$ and produces the best scores on 60\% classes of CIFAR-10 with an average AUC score of $0.67$.

\subsubsection{Protocol 2}
Table \ref{tab:protocol2MNIST-CIFAR} presents the results across the protocol 2 variant (1 normal/9 anomalous) across both MNIST and CIFAR-10. Our DAE+ALCN method obtains an AUC$_{avg}$ of 0.989 across MNIST and an AUC$_{avg}$ of 0.742 across CIFAR-10, outperforming all prior methods including OLED \cite{Jewell2022OLED} which uses discrete noise, as previously stated in this work. This gives illumination as to the benefit of using bespoke continuous noise while training.

\subsection{Real-world Tasks}

\begin{table}[htb!]
\centering
\caption{Quantitative results ($AUC_{avg}$) of models across Plant Village \cite{Hughes2015Plant_Village} dataset.}
\addtolength{\tabcolsep}{-4.2pt}
\begin{tabular}{|l|l|}
\hline

\multirow{2}{*}{\bf Model}  & {\bf Plant Village} \\ \cline{2-2} 
& {\bf $AUC_{avg}$} \\ \hline \hline
VAE \cite{Kingma2013} & 0.65 \\ \hline
AnoGAN \cite{Schlegl2017} & 0.65 \\ \hline
EGBAD \cite{zenati2018} & 0.70 \\ \hline
GANomaly \cite{Akcay2018} & 0.73 \\ \hline
Skip-GANomaly \cite{akccay2019skip} & 0.77 \\ \hline
\textbf{DAE+ALCN} & \textbf{0.77}\\ \hline

\end{tabular}
\addtolength{\tabcolsep}{0pt}  

\label{tab:plantvillage}
\end{table}

\subsubsection{MVTEC-AD Industrial Inspection Dataset}
\noindent In this experiment we compare our DAE+ALCN method against prior semi-supervised anomaly detection methods across the MVTEC-AD task \cite{bergmann2019mvtec} to verify that we can apply our method to a real-world example rather than solely across synthetic and trivial leave-one-out tasks. 

The results of this experiment are shown in Table \ref{tab:MVTEC}. It can be observed that DAE+ALCN obtains the highest average AUC score of $0.83$, outperforming all other methods on $10$ out of the $15$ classes in MVTEC-AD dataset.

\begin{table*}[htb!]
\centering
\caption{Comparison of model complexity (number of parameters (millions)) and inference time (milliseconds).}
\resizebox{\linewidth}{!}{
\begin{tabular}{ll|l|l|l|l|l|}
\cline{3-7}
& & \multicolumn{5}{c|}{\textbf{Model}} \\ \cline{3-7}
& & \textbf{DAE} & \textbf{AnoGAN }& \textbf{EGBAD} & \textbf{GANomaly} & \textbf{DAE + ALCN} \\ \cline{3-7} \hline 
\multicolumn{2}{|c|}{\textbf{Parameters (Million)}} & 1.12 & 233.04 & 8.65 & 3.86 & 9.87 \\ \hline \hline

\multicolumn{1}{|c|}{\textbf{Inference Time/Batch}} & \textbf{MNIST} & 2.36 & 667 & 8.02 & 9.7 & {\bf 4.54} \\ \cline{2-7} 
\multicolumn{1}{|c|}{\textbf{(Millisecond)}} & \textbf{CIFAR-10} & 2.73 & 611 & 9.55 & 10.53 & {\bf 5.23} \\ \hline

\end{tabular}
}
\addtolength{\tabcolsep}{0pt}  

\label{tab:complexity}

\end{table*}

    
    

\subsubsection{Plant Village Dataset} The Plant Village dataset \cite{bergmann2019mvtec} is challenging due to the large intra-class variance present in this dataset. Leaves of a given plant can vary vastly in appearance with respect to shape and colour. As such, it is challenging to map the underlying distribution of the leaves. The quantitative results of methods across this dataset are presented in Table \ref{tab:plantvillage}. Our DAE+ALCN method obtains an AUC$_{avg}$ of 0.77 which is the same as that of Skip-GANomaly \cite{akccay2019skip}. Both methods far-outperform prior methods across this dataset.

 \begin{figure}[htb!]
\centering
\includegraphics[width=\linewidth]{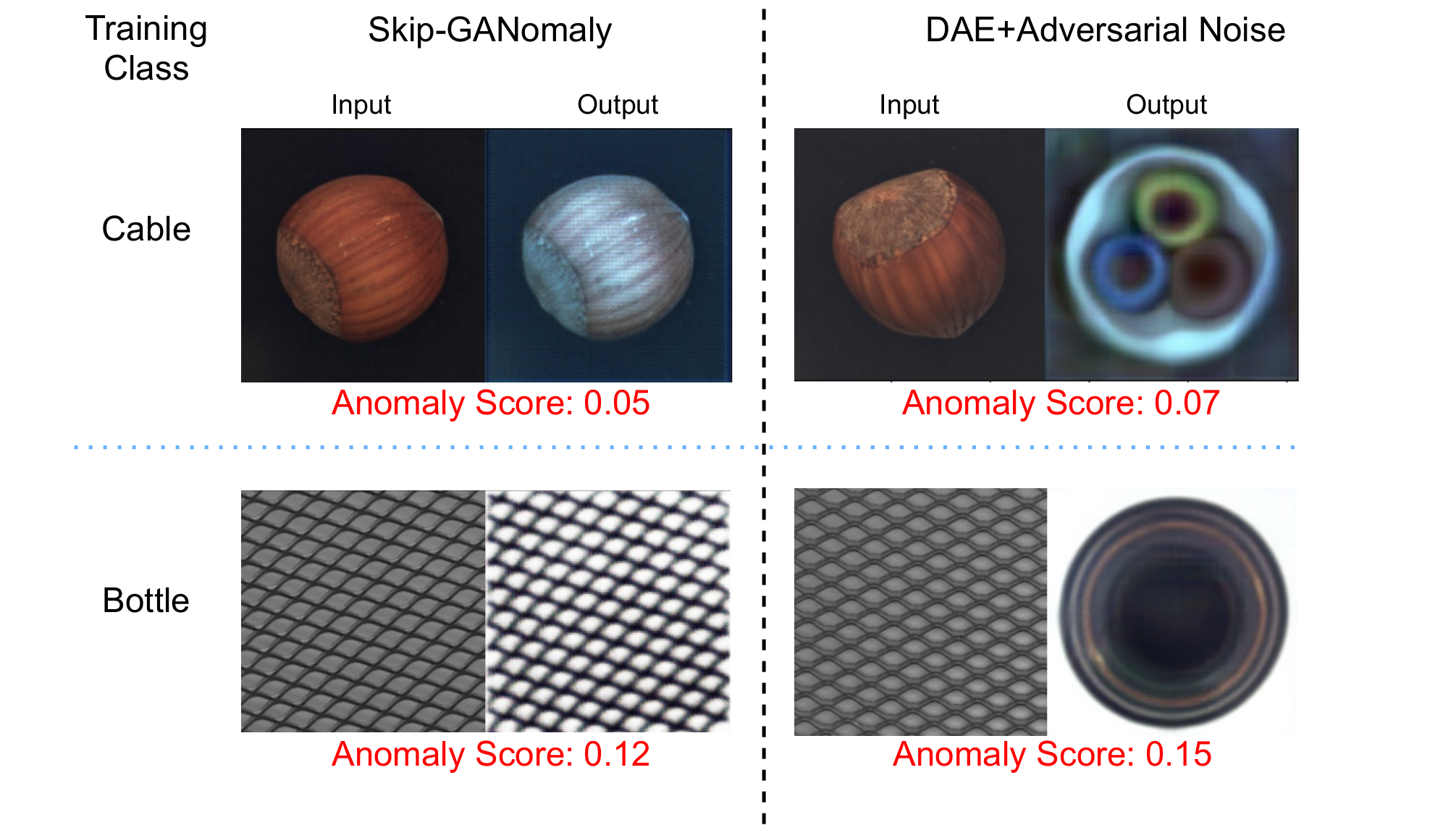}
\vspace{-0.6cm}
\caption{Comparison between Skip-GANomaly\cite{Akcay2019} and DAE+Adversarial Noise of feeding vastly out-of-distribution (Hazelnut and Grid) examples through models trained on a different class (Cable and Bottle). }
\label{fig:ToOthers}
\end{figure}

Figure \ref{fig:ToOthers} illustrates the results of an input which is an out-of-distribution example through both the Skip-GANomaly \cite{akccay2019skip} and DAE+ALCN into models trained on only another specific class singular (Figure \ref{fig:ToOthers}, left label). The objective being that Skip-GANomaly \cite{akccay2019skip} and DAE+ALCN should reconstruct out-of-distribution examples within the original class distribution. However, it can be seen in Figure \ref{fig:ToOthers} that Skip-GANomaly \cite{akccay2019skip} successfully reconstructs an out-of-distribution example given weight to the conclusion that it has converged to a pass-though identity function and just copies information from input to output (i.e. hazelnut/grid observed in both input + output), despite the fact the model has never been exposed to these class examples in training. For Skip-GANomaly this leads to low anomaly scores of 0.05 and 0.12 for Cable and Bottle respectively. By contrast, our DAE+ALCN architecture, manages to reconstruct such out-of-distribution examples back into the training classes thus resulting in the anomaly scores 0.07 for Cable (0.02 larger than Skip-GANomaly \cite{Akcay2019}) and 0.15 for Bottle (0.03 higher than Skip-GANomaly \cite{Akcay2019}). This shows that given vastly out-of-distribution examples, the DAE+ALCN network is more robust to misclassification and less prone to a pass-through identity-like reconstruction output.

Overall these experiments show that using our adversarial noise as a regularisation technique can enable even a simple architecture such as the Denoising Autoencoder outlined in Figure \ref{fig:model} to obtain better results than more complex model architectures. 

\subsection{Model Complexity}

\noindent An outline of model complexity together with inference time per batch is outlined in Table \ref{tab:complexity}. The DAE+ALCN architecture has $9.87$ Million parameters which is slightly larger than EGBAD \cite{zenati2018} which is at $8.65$ Million but still orders of magnitude smaller relative to that of AnoGAN \cite{Schlegl2017}. The magnitude of our model comes from the noise generation module (ALCN) in addition to the DAE module which is fairly light weight at 1.12 million parameters. This means that during training, the adversarial noising approach outlined in this paper adds a significant memory overhead to the model during training however, has an inference speed of $4$ milliseconds per batch which is significantly faster than the other methods, but generating the noise during inference adds a slight overhead of 2.5ms over the standard DAE architecture.   

\subsubsection{Qualitative Results}
\begin{figure*}[htb!]
\centering
\includegraphics[width=\linewidth]{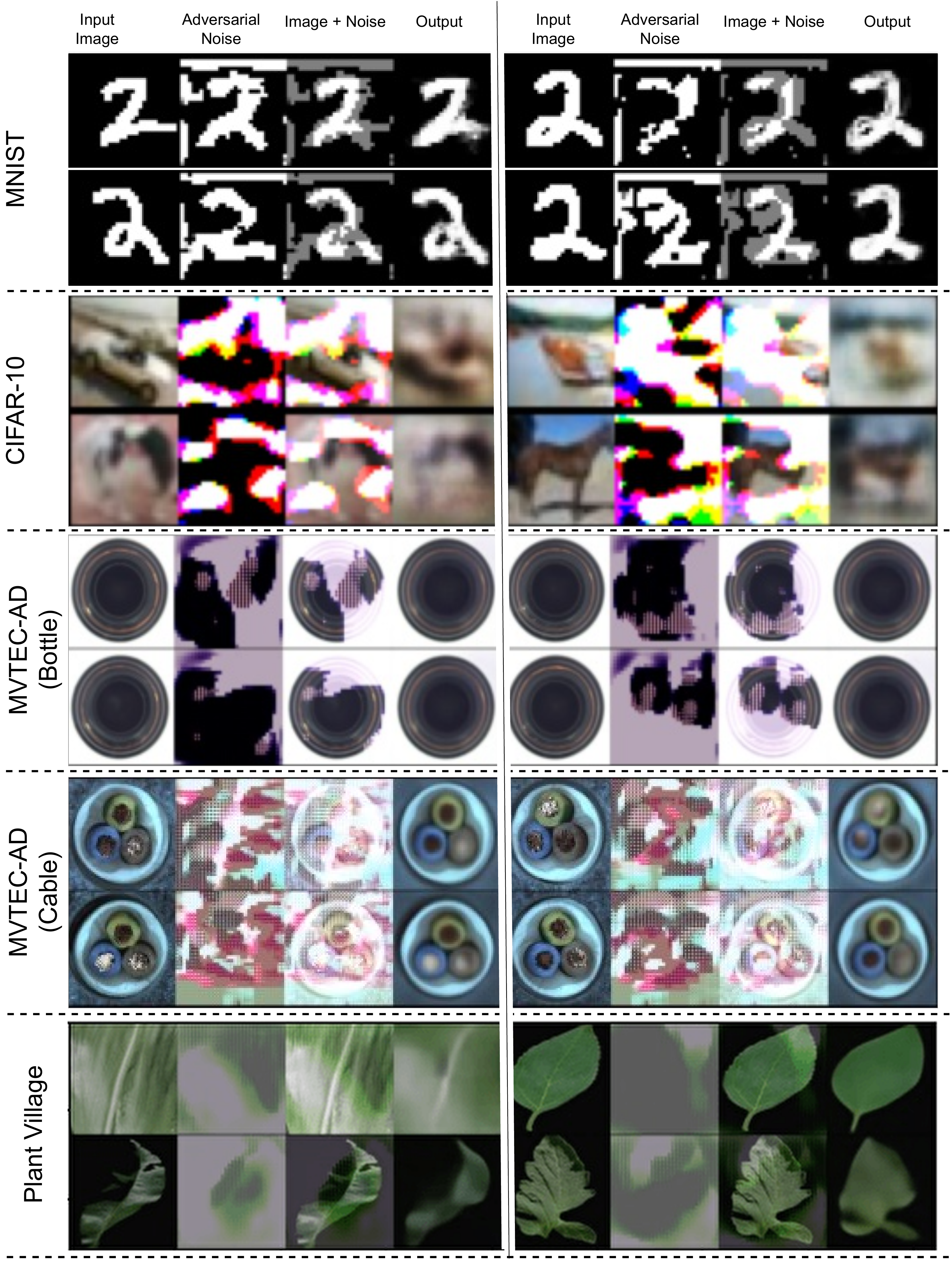}
\vspace{-0.6cm}
\caption{Examples of input image, generated adversarial noise, input + noise addition and resulting reconstructed output (left $\rightarrow$ right).}
\label{fig:Noise}
\end{figure*}

Figure \ref{fig:Noise} illustrates the qualitative results of DAE+ALCN across different datasets. The first column for each example shows the input images to the model. The second column illustrates the adversarial noise which is added to the input resulting in those images (3rd column). This adversarial noise + input is then fed into DAE and the resulting output after denoising (4th column). Of particular interest are the noise examples across the MNIST \cite{lecun-mnisthandwrittendigit-2010} and Plant Village \cite{Hughes2015Plant_Village} datasets. From Figure \ref{fig:Noise}  we can observe that the adversarial noise tends towards the style/shape of the input data which, when added to the image, adds a large level of input obfuscation (Figure \ref{fig:Noise} - Input + Noise columns). Despite this, the DAE architecture is able to successfully reconstruct the original input images from this maximally noised version with significant fidelity (Figure \ref{fig:Noise} - Output columns).

\section{Conclusion} \label{sec:conclusion}
\noindent In this work we introduce a novel approach for improved robustness semi-supervised anomaly detection by adversarially training a noise generator to produce maximal continuous noise which is then added to input data. In the same training step, a simple Denoising Autoencoder (DAE) is optimised to reconstruct the denoised, unperturbed input from the noised input. Through this simple approach, we vastly improve performance on semi-supervised anomaly detection tasks across both benchmark `leave-one-out' anomaly and challenging real-world anomaly detection tasks, outperforming prior work in the field. Via ablation, we also show the DAE with adversarial noise approach demonstrates superior performance against prior fixed-parameter noising strategies (random and Gaussian) across the leave-one-out benchmark tasks.


\begingroup
\setstretch{.94}
\bibliographystyle{apalike}
{\small
\bibliography{egbib}}
\endgroup


\end{document}